%% file: main.tex
\documentclass[12pt]{article}
\usepackage{booktabs}
\usepackage[a4paper, top=2.3cm, bottom=2.5cm, left=2cm, right=2cm]{geometry}% 12pt font size and single column
\usepackage[utf8]{inputenc}
\usepackage[T1]{fontenc}
\usepackage{lmodern}
\usepackage[hidelinks]{hyperref}
\usepackage{helvet}
\usepackage{graphicx}
\usepackage{amsmath, amssymb}
\usepackage{hyperref}
\usepackage{array}
\usepackage{tabularx}
\usepackage{enumitem}  % For custom item symbols
\usepackage{pifont}  
\usepackage{makecell} % For multi-line cells
\usepackage{longtable} % For tables spanning multiple pages
\usepackage{letterspace}
\usepackage{setspace}
\usepackage{titlesec} % If not already present
\usepackage{geometry} % If not already present
\usepackage{listings}
\usepackage{xcolor}
\usepackage{titling}
\usepackage{hyperref}
\usepackage{float} % Add to preamble if not already there
\usepackage{fancyhdr} % Page numbering
\usepackage{lipsum} 
\usepackage[most]{tcolorbox}
\usepackage{parskip}

\usepackage{fvextra}  % Better verbatim handling

\definecolor{jsonstring}{RGB}{173, 0, 0}   % Red for strings
\definecolor{jsonkey}{RGB}{0, 150, 0}      % Blue for keys
\definecolor{jsonnumber}{RGB}{150, 0, 150} % Magenta for numbers
\definecolor{jsonbg}{RGB}{240, 240, 240}   % Light gray background

\newtcolorbox{jsonbox}{
    breakable,        % Makes the box split across pages
    colback=jsonbg,
    colframe=black,
    boxrule=0.5pt,
    arc=2mm,
    fontupper=\ttfamily\footnotesize,
    before upper={\setlength{\parindent}{0pt}},
}

\newcommand{\jsonkey}[1]{\textcolor{jsonkey}{"#1"}}
\newcommand{\jsonstring}[1]{\textcolor{jsonstring}{"#1"}}
\newcommand{\jsonnumber}[1]{\textcolor{jsonnumber}{$#1$}}

\definecolor{lightgray}{rgb}{0.9,0.9,0.9}
\begin{document}
\titlespacing*{\section}{0pt}{0pt}{-0.7cm} % Space below section titles (already present)
\titlespacing*{\subsection}{-10pt}{0pt}{-0.4cm} % Indent subsections by 15pt, space below 0.3cm
\titlespacing*{\subsubsection}{-10pt}{0pt}{-0.4cm} % Indent subsubsections by 25pt, space below 0.2cm
% Single-column is default in 'journal' mode, no need to force it
% \IEEEoverridecommandlockouts - Not needed for single-column journal style
\renewcommand{\familydefault}{\sfdefault}
\renewcommand{\abstractname}{} % Remove "Abstract" title

% Title
\input{title.tex}

% Abstract
\input{abstract.tex}
\vspace{0cm} % Add negative vertical space after abstract (adjust value as needed)
\clearpage
\section *{}
\input{toc.tex}
% Sections - Keep \section*{} for unnumbered sections
\clearpage
\section *{}
\input{introduction.tex}
\clearpage
\section *{}
\input{methods.tex}
\clearpage
\section*{}
\input{results.tex}
\clearpage
\section*{}
\input{discussion.tex}
\clearpage
\section*{}
\input{conclusion.tex}

% References

\clearpage
\section*{} % Unnumbered References section

\input{references.tex}
\clearpage

% Appendix
\section*{}
\input{appendix.tex}
\end{document}

%% file: title.tex
\label{title}
\begin{center}
\begin{center}
\begin{center}
\begin{spacing}{1.3}
    {\fontsize{17pt}{20pt}\selectfont\textbf{The Reliability of LLMs for Medical Diagnosis: \\ An Examination of Consistency, Manipulation, and Contextual Awareness}} % Lineskip set to 54pt for "lineheight of 3"
\end{spacing}
\end{center}

\end{center}

    \vspace{0.4cm} % Adjust vertical space after title
    \large
    {\fontsize{12pt}{14.4pt}\selectfont Krishna Subedi  \par} % Author name in a more balanced size

    \vspace{0.1cm} % Small space between author and email

    {\fontsize{11pt}{12pt}\selectfont \texttt{krishna.subedi@neryva.com}} % Email in a smaller, yet readable font
    \vspace{0.5cm}
\end{center}

%% file: abstract.tex
\label{abstract}
\begin{center}

\textbf{Abstract}
\end{center}
\begin{spacing}{1.2}
\vspace{0.2cm}
This study evaluated the diagnostic reliability of two Large Language Models (LLMs), Google Gemini 2.0 Flash and OpenAI ChatGPT-4o, across three dimensions: consistency under rephrased inputs, susceptibility to irrelevant prompt content, and responsiveness to added clinical context~\cite{npjDigitalMedicine2024, Berg2024, Amodei2016, Rajpurkar2022}.

We designed 52 clinical scenarios and modified each under controlled conditions. For consistency, scenarios were rephrased with demographic, wording, and examination changes that preserved the diagnostic core. And the susceptibility was evaluated through embedding irrelevant but plausible narrative details while keeping the clinical evidence unchanged. For contextual awareness, patient history, lifestyle data, or diagnostic findings were added to shift the expected diagnosis. Physician reviewers then judged whether context-driven changes were clinically appropriate.

Both models returned identical diagnoses across all equivalent variants and repeated queries (\textbf{100\% consistency}). When irrelevant details were added, Gemini changed its diagnosis in \textbf{40.0\%} of cases and ChatGPT in \textbf{30.0\%}. ChatGPT responded to context more often than Gemini (\textbf{77.8\%} vs. \textbf{55.6\%}), but a larger share of its changes were clinically inappropriate (\textbf{33.3\%} vs. \textbf{22.2\%}). Gemini's context-driven changes were more often judged appropriate (\textbf{66.7\%} vs. \textbf{55.6\%}).

Consistency under controlled inputs did not protect either model from irrelevant manipulation or unjustified diagnostic shifts when context changed. Before LLMs can separate relevant from irrelevant input and flag insufficient evidence, their diagnostic use requires clinician oversight and structured safeguards.
\end{spacing}

%% file: toc.tex
% Custom Table of Contents Design
\section*{\centering \textbf{Table of Contents}}
\vspace{1.6cm}

% Main Sections with added vertical spacing
\noindent \hyperref[title]{\textbf{Title}} \dotfill \pageref{title} \par
\vspace{0.5em}
\noindent \hyperref[abstract]{\textbf{Abstract}} \dotfill \pageref{abstract} \par
\vspace{0.5em}
\noindent \hyperref[sec:intro]{\textbf{I. Introduction}} \dotfill \pageref{sec:intro} \par
\vspace{0.5em}
\noindent \hyperref[sec:methods]{\textbf{II. Method}} \dotfill \pageref{sec:methods} \par
\vspace{0.5em}
\hspace{1em} \hyperref[sec:study_design]{2.1. Study Design} \dotfill \pageref{sec:study_design} \par
\vspace{0.3em}
\hspace{1em} \hyperref[sec:primary_objectives]{2.2. Primary Objectives} \dotfill \pageref{sec:primary_objectives} \par
\vspace{0.3em}
\hspace{1em} \hyperref[sec:llm_selection]{2.3. LLM Selection and Configuration} \dotfill \pageref{sec:llm_selection} \par
\vspace{0.3em}
\hspace{1em} \hyperref[sec:clinical_scenarios]{2.4. Clinical Scenarios} \dotfill \pageref{sec:clinical_scenarios} \par
\vspace{0.3em}
\hspace{1em} \hyperref[sec:consistency_assessment]{2.5. Consistency Assessment Protocol} \dotfill \pageref{sec:consistency_assessment} \par
\vspace{0.3em}
\hspace{1em} \hyperref[sec:susceptibility_manipulation]{2.6. Susceptibility to Manipulation Protocol} \dotfill \pageref{sec:susceptibility_manipulation} \par
\vspace{0.3em}
\hspace{1em} \hyperref[sec:contextual_awareness]{2.7. Contextual Awareness Assessment Protocol} \dotfill \pageref{sec:contextual_awareness} \par
\vspace{0.3em}
\hspace{1em} \hyperref[sec:data_analysis]{2.8. Data Analysis} \dotfill \pageref{sec:data_analysis} \par
\vspace{0.3em}
\hspace{1em} \hyperref[sec:ethical_considerations]{2.9. Ethical Considerations and Limitations} \dotfill \pageref{sec:ethical_considerations} \par
\vspace{0.5em}
\noindent \hyperref[sec:results]{\textbf{III. Results}} \dotfill \pageref{sec:results} \par
\vspace{0.5em}
\hspace{1em} \hyperref[sec:diagnostic_consistency]{3.1. Diagnostic Consistency} \dotfill \pageref{sec:diagnostic_consistency} \par
\vspace{0.3em}
\hspace{1em} \hyperref[sec:susceptibility_results]{3.2. Susceptibility to Manipulation} \dotfill \pageref{sec:susceptibility_results} \par
\vspace{0.3em}
\hspace{1em} \hyperref[sec:contextual_awareness_results]{3.3. Contextual Awareness} \dotfill \pageref{sec:contextual_awareness_results} \par
\vspace{0.5em}
\noindent \hyperref[sec:discussion]{\textbf{IV. Discussion}} \dotfill \pageref{sec:discussion} \par
\vspace{0.5em}
\hspace{1em} \hyperref[sec:discussion_consistency]{4.1. Diagnostic Consistency} \dotfill \pageref{sec:discussion_consistency} \par
\vspace{0.3em}
\hspace{1em} \hyperref[sec:discussion_susceptibility]{4.2. Susceptibility to Manipulation} \dotfill \pageref{sec:discussion_susceptibility} \par
\vspace{0.3em}
\hspace{1em} \hyperref[sec:discussion_contextual]{4.3. Contextual Awareness} \dotfill \pageref{sec:discussion_contextual} \par
\vspace{0.3em}
\hspace{1em} \hyperref[sec:integrated_discussion]{4.4. Synthesis} \dotfill \pageref{sec:integrated_discussion} \par
\vspace{0.3em}
\hspace{1em} \hyperref[sec:implications_clinical]{4.5. Implications for Clinical Use} \dotfill \pageref{sec:implications_clinical} \par
\vspace{0.3em}
\hspace{1em} \hyperref[sec:limitations_future]{4.6. Limitations and Future Work} \dotfill \pageref{sec:limitations_future} \par
\vspace{0.7em}
\noindent \hyperref[sec:conclusion1]{\textbf{V. Conclusion}} \dotfill \pageref{sec:conclusion1} \par
\vspace{0.5em}
\noindent \hyperref[sec:references]{\textbf{REFERENCES}} \dotfill \pageref{sec:references} \par
\vspace{0.5em}
\noindent \hyperref[sec:app1]{\textbf{APPENDIX A: Data Availability}} \dotfill \pageref{sec:app1} \par
\vspace{0.3em}
\noindent \hyperref[sec:app2]{\textbf{APPENDIX B: Example Baseline Scenario}} \dotfill \pageref{sec:app2} \par
\vspace{0.3em}
\noindent \hyperref[sec:app3]{\textbf{APPENDIX C: Prompt Sample}} \dotfill \pageref{sec:app3} \par

%% file: introduction.tex
\textbf{I. Introduction} \vspace{-10pt} \\
\phantomsection
\label{sec:intro}
\begin{spacing}{1.2}

Access to timely and accurate diagnosis remains uneven, and the gap is widest where specialists and diagnostic infrastructure are scarce. Large Language Models (LLMs)~\cite{Vaswani2017, Devlin2018, Radford2019, Brown2020} have been proposed for clinical reasoning, triage, education, and documentation~\cite{npjDigitalMedicine2024, Berg2024, Bubeck2023}. LLMs can parse natural-language case descriptions and produce plausible diagnostic outputs~\cite{Rajpurkar2022, Topol2019}, but clinical usefulness demands more than plausible answers under ideal conditions.

Most evaluations in medical LLMs report diagnostic accuracy on benchmark cases~\cite{Esteva2019, Rajpurkar2022}. Accuracy matters, but it does not show whether a model is reliable enough for clinical use~\cite{Amodei2016, Marcus2020}. A diagnostic system should also produce stable outputs when the same clinical information is presented in equivalent forms, resist irrelevant or misleading input, and weigh contextual information against the clinical evidence~\cite{Finlayson2019, Goodfellow2015, Szegedy2014}. Real clinical narratives are rarely clean: patients include irrelevant details, omit findings, describe symptoms imprecisely, or introduce context that shifts the interpretation~\cite{Ghassemi2021, Mittelstadt2016}.

This study tests three reliability dimensions. \textbf{Diagnostic consistency}: does the LLM give clinically equivalent diagnoses when the same case is repeated or superficially varied? \textbf{Susceptibility to manipulation}: can irrelevant but plausible prompt content shift the model away from the diagnosis supported by the clinical evidence? \textbf{Contextual awareness}: does the model incorporate patient history, lifestyle factors, and diagnostic findings without overreacting to weak or misleading cues?

LLMs do not evaluate information the way clinicians do~\cite{Shortliffe2018, Marcus2020}. A clinician questions inconsistent data, asks follow-up questions, judges whether a provided detail by the patient is relevant or not, and defers diagnosis when required information is missing. Current LLMs tend to treat all prompt content as usable evidence and produce a final diagnosis even when the input is incomplete or noisy~\cite{Brown2020, OpenAIResearchTeam2023, Floridi2020}. The risk grows when LLMs are used by non-specialists, patients, or clinicians in settings where independent verification is difficult~\cite{Floridi2018, Floridi2019, Char2018}.
	\vspace{0.5cm}
	This study addresses the following research questions:

\begin{itemize}
    \item How consistent are LLM diagnostic recommendations when clinically equivalent scenarios are repeated or presented with minor demographic, wording, or examination-phrasing variations?
    \item How often do irrelevant but plausible additions to a prompt change an LLM's diagnosis when the core clinical evidence remains unchanged?
    \item How appropriately do LLMs incorporate contextual information (patient history, lifestyle factors, medication history, and diagnostic findings) when generating diagnoses?

\end{itemize}
To answer these questions, this study evaluates Gemini and ChatGPT using controlled clinical scenarios, systematic prompt variations, manipulation tests, and physician review of context-driven diagnostic changes.

\end{spacing}

%% file: methods.tex
\textbf{II. Method} \vspace{-10pt} \\
\phantomsection
\label{sec:methods}

\begin{spacing}{1.2}
This study used a controlled comparative design to evaluate the diagnostic reliability of Large Language Models (LLMs) across three dimensions: diagnostic consistency, susceptibility to manipulation, and contextual awareness.

\subsection*{}
\textbf{2.1. Study Design}
\phantomsection
\label{sec:study_design}
\par
Gemini and ChatGPT received systematically modified clinical prompts under three conditions. Clinically equivalent case variants tested consistency. Irrelevant but plausible narrative additions, with the core evidence unchanged, tested susceptibility to manipulation. Clinically relevant context expected to shift the diagnosis tested contextual awareness. Outputs were compared quantitatively and, where clinical judgment was needed, reviewed by physicians.

\subsection*{} % 2.2. Primary Objectives
\textbf{2.2. Primary Objectives}
\phantomsection
\label{sec:primary_objectives}
\par
The objectives were to:
\begin{itemize}
    \item \textbf{Quantify diagnostic consistency} across repeated presentations and clinically equivalent case variants.
    \item \textbf{Measure susceptibility to manipulation} by determining whether irrelevant prompt additions changed model diagnoses.
    \item \textbf{Evaluate contextual awareness} by assessing whether clinically relevant contextual changes produced appropriate diagnostic shifts.
    \item \textbf{Compare model performance} between Gemini and ChatGPT across these reliability dimensions.
    \item \textbf{Assess clinical appropriateness} of diagnostic changes through physician review.
\end{itemize}

\subsection*{} % 2.3. LLM Selection and Configuration
\textbf{2.3. LLM Selection and Configuration}
\phantomsection
\label{sec:llm_selection}

Two commercially available LLMs were evaluated:
\begin{itemize}
    \item \textbf{Google Gemini 2.0 Flash} (accessed between February 2 and February 7, 2025)
    \item \textbf{OpenAI ChatGPT-4o} (accessed between February 2 and February 7, 2025)~\cite{OpenAIResearchTeam2023}
\end{itemize}
Both models were accessed through their public chat interfaces, reflecting how clinicians, healthcare workers, or patients would use them for diagnostic queries~\cite{Bubeck2023}.

\subsection*{} % 2.4. Clinical Scenarios
\textbf{2.4. Clinical Scenarios}
\phantomsection
\label{sec:clinical_scenarios}
\par
The dataset contained \textbf{52 de novo clinical scenarios} representing \textbf{39 unique medical conditions}. Each scenario supported controlled modification of patient details, symptoms, examination findings, test results, and contextual factors.

\subsubsection*{} % 2.4.1. Scenario Characteristics:
\textbf{2.4.1. Scenario Characteristics}

The conditions covered eight broad clinical areas:
\begin{itemize}
    \item \textbf{Cardiovascular diseases:} myocardial infarction, angina pectoris, hypertension, and heart failure.
    \item \textbf{Pulmonary diseases:} acute bronchitis, COPD, asthma, COVID-19, and influenza.
    \item \textbf{Neurological diseases:} ischemic stroke, transient ischemic attack, subarachnoid haemorrhage, migraine, diabetic neuropathy, and generalised anxiety disorder.
    \item \textbf{Endocrine and metabolic disorders:} type 1 diabetes mellitus, type 2 diabetes mellitus, and hypothyroidism.
    \item \textbf{Gastrointestinal diseases:} peptic ulcer disease, appendicitis, diverticulitis, viral gastroenteritis, and constipation.
    \item \textbf{Musculoskeletal diseases:} lumbar strain, osteoarthritis, rheumatoid arthritis, and distal radius fracture.
    \item \textbf{Infectious diseases:} urinary tract infection, common cold, streptococcal pharyngitis, viral exanthem, and acute otitis media.
    \item \textbf{Pain and miscellaneous presentations:} musculoskeletal pain and related nonspecific complaints.
\end{itemize}
Scenario complexity ranged from common outpatient presentations to acute or diagnostically complex cases such as STEMI, subarachnoid haemorrhage, COPD exacerbation, heart failure subtypes, and urinary tract infection subtypes.

\subsubsection*{} % 2.4.2. Dataset Construction and Structure:
\textbf{2.4.2. Dataset Construction and Structure}
\par
UpToDate and DynaMed served as primary sources for symptoms, vital signs, diagnostic criteria, risk factors, differential diagnoses, final diagnoses, and treatment recommendations. ClinicalKey and PubMed/MEDLINE provided supplementary detail where needed.

Each scenario used a standardised structure:
\begin{itemize}
    \item \textbf{Patient information:} \texttt{patient\_id}, age, and gender.
    \item \textbf{Medical background:} medical history and current medications.
    \item \textbf{Presenting complaint:} the primary reason for presentation.
    \item \textbf{Symptoms:} symptom name, severity, character, associated symptoms, exacerbating factors, relieving factors, and symptom type where applicable.
    \item \textbf{Vital signs:} heart rate, blood pressure, temperature, and respiratory rate.
    \item \textbf{Physical examination:} pertinent examination findings.
    \item \textbf{Diagnostic test results:} relevant ECG, troponin, complete blood count, chest X-ray, laboratory, imaging, or condition-specific findings.
    \item \textbf{Differential diagnosis:} candidate diagnoses considered during case construction.
    \item \textbf{Final diagnosis:} the expected or most likely diagnosis.
    \item \textbf{Treatment:} guideline-aligned management recommendations.
    \item \textbf{Additional notes:} relevant contextual or follow-up details.
\end{itemize}

\subsubsection*{} % 2.4.3.Data Modification and Open Access:
\textbf{2.4.3. Data Modification and Open Access}

Non-clinical fields (\texttt{patient\_id}, age, gender, medical history wording, and presenting complaint phrasing) were modified by hand for each experimental condition. All modifications stayed clinically plausible and, unless the condition required otherwise, preserved the diagnostic core of the case.

\begin{center}
    \begin{figure}[htbp]
    \centering
    \includegraphics[width=0.99\textwidth]{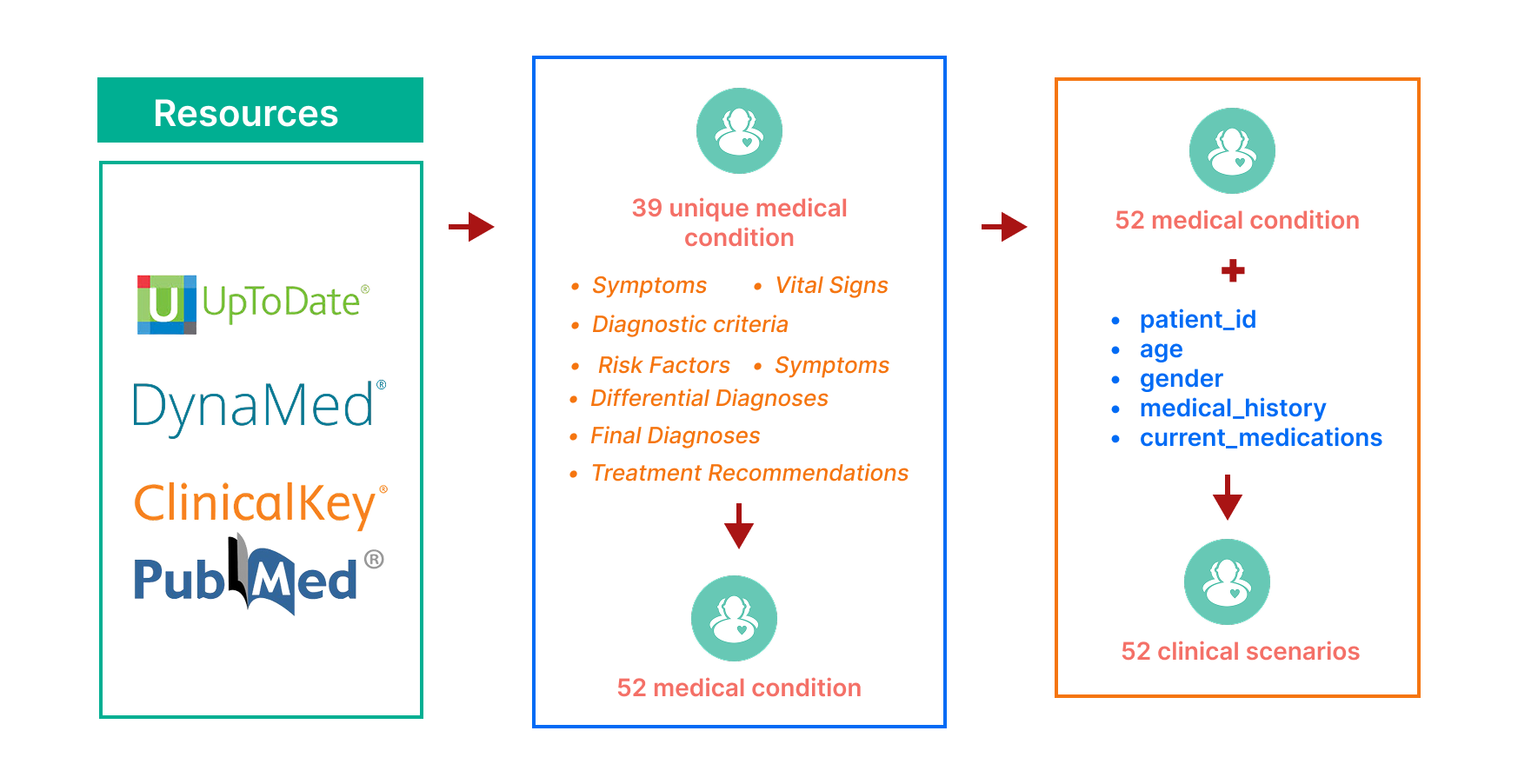}
    \caption{\fontsize{10pt}{20pt}\textit{Dataset Construction Methodology}}
    \label{fig:dataset_methodology}
\end{figure}
\end{center}
\vspace{-0.9cm}
The 52 clinical scenarios are available in the public repository listed in Appendix A.

\subsubsection*{} % 2.5. Consistency Assessment Protocol
\textbf{2.5. Consistency Assessment Protocol}
\phantomsection
\label{sec:consistency_assessment}
\par
Each model received the same case multiple times with superficial variation to test whether it returned clinically equivalent diagnoses.

\subsubsection*{} % 2.5.1 Clinical Scenario Variations for Consistency Testing:
\textbf{2.5.1. Clinical Scenario Variations for Consistency Testing} \par
Four clinically equivalent variants were curated for each of the 52 baseline scenarios by changing superficial details in three categories while preserving the diagnostic core:
\begin{itemize}
    \item \textbf{Demographics:} country, race, ethnicity, age, and gender were varied when those changes were not intended to alter the expected diagnosis.
    \item \textbf{Presenting complaint phrasing:} symptom descriptions were reworded using clinically equivalent language.
    \item \textbf{Medical history and physical examination phrasing:} descriptive wording was revised without changing the underlying clinical findings.
\end{itemize}

\subsubsection*{} % 2.5.2. Standardized Prompt Structure and Procedure:
\textbf{2.5.2. Standardised Prompt Structure and Procedure}

All consistency prompts used the following format:
\begin{quote}
\textbf{You are a medical doctor. Diagnose the following patient case:}

\texttt{[Patient Data]}

\textbf{Provide \texttt{patient\_id} and the corresponding diagnosis for that patient. Provide \textbf{only} your final diagnosis. Do not include explanations or differential diagnoses.}
\end{quote}

Each model received 624 consistency queries (52 scenarios $\times$ 4 variants $\times$ 3 repetitions), and all outputs were recorded.

\subsubsection*{} % 2.5.3. Criteria for Determining Diagnostic Consistency:
\textbf{2.5.3. Criteria for Determining Diagnostic Consistency}

Diagnostic consistency was evaluated at two levels:
\begin{itemize}
    \item \textbf{Internal variation consistency:} the three repeated outputs for a given variant were classified as internally consistent if they were identical or clinically equivalent.
    \item \textbf{Baseline scenario consistency:} a baseline scenario was classified as consistent for a model only if all four variants produced clinically equivalent diagnoses across repetitions.
\end{itemize}
Diagnoses were clinically equivalent if they matched exactly or used synonymous terms, for example, \textit{gastroesophageal reflux disease} and \textit{acid reflux}. Any clinically distinct diagnosis in any variant or repetition counted as inconsistency.

\subsection*{} % 2.6. Susceptibility to Manipulation Protocol
\textbf{2.6. Susceptibility to Manipulation Protocol}
\phantomsection
\label{sec:susceptibility_manipulation}
\par
Here, the question was whether irrelevant but plausible prompt additions could shift a model's diagnosis when the core clinical evidence remained unchanged.

\subsubsection*{} % 2.6.1. Manipulation Strategy
\textbf{2.6.1. Manipulation Strategy}

Ten clinical scenarios were randomly selected. For each, a manipulated counterpart embedded irrelevant narrative details while preserving all symptoms, history, vital signs, physical findings, and test results needed for the original diagnosis. The irrelevant additions fell into six categories.

\begin{longtable}{>{\raggedright\arraybackslash}p{4.0cm} >{\raggedright\arraybackslash}p{4.5cm} p{6.5cm}}
\caption{Examples of manipulation strategies and irrelevant information introduced in diagnostic scenarios.}\label{tab:manipulation_strategies} \\
\toprule
\textbf{Category} & \textbf{Manipulation Strategy} & \textbf{Examples of Introduced Irrelevant Information} \\
\midrule
\endfirsthead
\caption[]{(Continued)} \\
\toprule
\textbf{Category} & \textbf{Manipulation Strategy} & \textbf{Examples of Introduced Irrelevant Information} \\
\midrule
\endhead
\textbf{Whimsical/\newline Figurative Language} & Use of metaphorical, humorous, or exaggerated symptom descriptions. & \textbf{Abdominal pain rated 9/10, like stepping on a Lego barefoot}; \textbf{fatigue feels like wading through treacle}; \textbf{chest pain described as a balloon popping in my chest}. \\[6pt]
\addlinespace
\addlinespace
\textbf{Alternative Medicine Beliefs} & References to unproven therapies or dismissive attitudes toward conventional medicine. & \textbf{Patient insists apple cider vinegar is a universal cure}; \textbf{prefers chakra alignment therapy over conventional pain relievers}; \textbf{believes high cholesterol is a modern myth}. \\[6pt]
\addlinespace
\addlinespace
\textbf{Anecdotal/\newline Subjective Claims} & Inclusion of unverified patient intuitions, personal theories, or irrelevant boasts. & \textbf{Claims to predict rain based on joint pain}; \textbf{boasts of being healthy as a horse}; \textbf{attributes mood changes to the aurora borealis seen online}. \\[6pt]
\addlinespace
\addlinespace
\textbf{Cultural/\newline Regional References} & Location-specific idioms, habits, or travel history not relevant to diagnosis. & \textbf{Uses British idioms such as get-up-and-go just got up and went}; \textbf{mentioned returning from a yoga retreat in Sedona}; \textbf{observed wearing a woolly scarf indoors despite mild temperatures}. \\[6pt]
\addlinespace
\addlinespace
\textbf{Patient\newline Demeanor/\newline Psychology} & Descriptions of emotional states, personality traits, or consultation behaviours. & \textbf{Sighed wistfully throughout the consultation}; \textbf{more concerned about missing a yoga class than chest pain}; \textbf{used excessive air quotes and wellness jargon}. \\[6pt]
\addlinespace
\addlinespace
\textbf{Irrelevant Lifestyle Details} & Unrelated hobbies, family history, or personal preferences. & \textbf{Collects vintage teacups as a hobby}; \textbf{father is an avid vinyl record collector}; \textbf{takes ginger supplements daily for an energy boost}; \textbf{enjoys knitting and watching period dramas}. \\
\bottomrule
\end{longtable}

\subsubsection*{} % 2.6.2. Prompt Structure and Susceptibility Testing Procedure:
\textbf{2.6.2. Prompt Structure and Susceptibility Testing Procedure}

Original and manipulated versions of each scenario were presented once to each model using the diagnostic prompt from Section 2.5.2. Outputs were compared within each pair.

\subsubsection*{} % 2.6.3. Criteria for Determining Susceptibility to Manipulation:
\textbf{2.6.3. Criteria for Determining Susceptibility to Manipulation}

A model was classified as susceptible for a scenario if the manipulated prompt produced a clinically distinct diagnosis, one representing a different medical condition or one that would alter clinical interpretation or management.

\subsection*{} % 2.7 Contextual Awareness Assessment Protocol
\textbf{2.7. Contextual Awareness Assessment Protocol}
\phantomsection
\label{sec:contextual_awareness}
\par
Unlike the manipulation test, which added irrelevant information, this phase added clinically relevant context and asked whether models changed their diagnoses in response.

\subsubsection*{} % 2.7.1 Approach to Scenario Modification:
\textbf{2.7.1. Approach to Scenario Modification}

Two baseline scenarios were randomly selected. One was modified into four context-rich versions and the second into five, producing \textbf{nine contextually varied scenarios}. The contextual modifications included:
\begin{itemize}
    \item \textbf{Demographics:} age, gender, race/ethnicity, and country of origin.
    \item \textbf{Clinical presentation:} presenting complaint, symptom severity, duration, location, character, associated symptoms, and relieving or aggravating factors.
    \item \textbf{Physical examination:} findings relevant to the intended diagnostic shift.
    \item \textbf{Medical history and medications:} additions, removals, or revisions to conditions and medication use.
    \item \textbf{Diagnostic test results:} ECG, laboratory, imaging, or other test findings aligned with the modified clinical presentation.
\end{itemize}

\subsubsection*{} % 2.7.2 Standardized Prompt Structure:
\textbf{2.7.2. Standardised Prompt Structure}

The same diagnostic prompt from Section 2.5.2 was used.

\subsubsection*{} % 2.7.3. Context-Rich and Context-Absent Scenarios:
\textbf{2.7.3. Context-Rich and Context-Absent Scenarios}

Context-rich scenarios contained clinically relevant patient details that could alter diagnostic interpretation; context-absent prompts used sparse clinical descriptions.
\begin{table*}[ht]
\centering
\renewcommand{\arraystretch}{1.5}
\begin{tabular}{p{5cm} p{5.5cm} p{5cm}}
\toprule
\textbf{Context Layer} & \textbf{Context-Rich Example} & \textbf{Context-Absent Example} \\
\midrule
\textbf{Demographics} & 35-year-old South Asian male & Adult patient \\
\textbf{Medical History} & History of hypertension, GERD, currently on pantoprazole & No significant past medical history reported \\
\textbf{Symptoms} & Epigastric pain radiating to back, worsens after spicy meals & Abdominal pain \\
\textbf{Lifestyle/Geographic} & Current smoker, works night shifts in a factory setting & \textit{[Omitted: no lifestyle/geographic details]} \\
\textbf{Diagnostic Test Results} & H. pylori stool antigen test is positive & Routine lab results are within normal limits \\
\bottomrule
\end{tabular}
\caption{Examples of Contextual Layers and Variations}
\label{tab:contextual_layers}
\end{table*}

\subsubsection*{} % 2.7.4. Illustrative Scenario Modifications:
\textbf{2.7.4. Illustrative Scenario Modifications} \par
For illustration, a sparse chest-pain case could be shifted in two directions: toward a lower-risk non-cardiac presentation by changing the patient to a younger woman with anxiety disorder, GERD history, upper back discomfort after emotional stress, normal ECG, and positive H. pylori testing; or toward a higher-risk cardiac presentation by changing the patient to an older man with diabetes, hypertension, smoking history, crushing substernal chest pain, ST depression, and elevated LDL.

\subsubsection*{} % 2.7.5. Example Prompts: Context-Absent vs. Context-Rich
\textbf{2.7.5. Example Prompts: Context-Absent vs. Context-Rich}

Table 3 shows illustrative prompt formats; these do not represent the full dataset.
\begin{table}[ht]
\centering
\renewcommand{\arraystretch}{1.5}
\begin{tabularx}{\textwidth}{p{2.5cm} p{2.5cm} X}
\toprule
\textbf{Case} & \textbf{Variation Type} & \textbf{Prompt Example} \\
\midrule
\textbf{Case 1} & \textbf{Context-Absent} &
\textit{You are a medical doctor. Diagnose the following patient case:}
A patient presents with chest discomfort. ECG abnormal.
Provide diagnosis. Provide ONLY your final diagnosis.
Do not provide explanations. \\

\textbf{Case 1} & \textbf{Context-Rich} &
\textit{You are a medical doctor. Diagnose the following patient case:}
A 28-year-old female with no cardiac history reports sharp left-sided chest pain worsening with deep inspiration.
Recent travel to a high-altitude region. D-dimer 0.3 micrograms/mL. ECG normal.
Diagnose. Provide ONLY your final diagnosis. Do not provide explanations. \\

\textbf{Case 2} & \textbf{Context-Absent} &
\textit{You are a medical doctor. Diagnose the following patient case:}
Patient complains of abdominal pain and nausea.
Endoscopy shows an ulcer.
Provide diagnosis. Provide ONLY your final diagnosis.
Do not provide explanations. \\

\textbf{Case 2} & \textbf{Context-Rich} &
\textit{You are a medical doctor. Diagnose the following patient case:}
A 65-year-old Japanese male with daily NSAID use for osteoarthritis presents with melena and epigastric tenderness.
H. pylori negative. Hemoglobin 9.2 g/dL.
Diagnose. Provide ONLY your final diagnosis. Do not provide explanations. \\
\bottomrule
\end{tabularx}
\caption{Comparison of context-absent and context-rich variations in diagnostic prompts.}
\label{tab:context_variations}
\end{table}

\subsubsection*{} % 2.7.6 Methodology for Diagnosis Comparison and Contextual Awareness Scoring
\textbf{2.7.6. Methodology for Diagnosis Comparison and Contextual Awareness Scoring}

For each context-rich scenario, the model's diagnosis was compared with the expected diagnosis defined during scenario construction. A match required exact agreement or clinical equivalence as defined in Section 2.5.3.

\begin{equation}
    \text{Diagnostic Match Rate} = \left( \frac{\text{Number of Matched Diagnoses}}{\text{Total Number of Contextually Varied Scenarios}} \right) \times 100\%
\end{equation}

\subsubsection*{} % 2.7.7. Qualitative Review of Contextual Appropriateness
\textbf{2.7.7. Qualitative Review of Contextual Appropriateness}

Two independent board-certified physicians reviewed all context-driven diagnostic changes. Each reviewer received the context-absent baseline, each context-rich variation, the diagnoses from both models, and the expected diagnosis. Changes were categorised as:
\begin{itemize}
    \item \textbf{Appropriate change:} clinically justified, evidence-based, and aligned with the intended contextual shift.
    \item \textbf{Inappropriate change:} clinically unjustified, erroneous, illogical, or based on misinterpretation of the contextual information.
    \item \textbf{Ambiguous change:} not clearly appropriate or inappropriate based on the available information, or requiring additional clinical data.
\end{itemize}
Inter-rater agreement, measured by Cohen's kappa, was $\kappa = 0.85$. Disagreements were resolved through discussion with a third senior clinician.

\subsection*{} % 2.8. Data Analysis
\textbf{2.8. Data Analysis}
\phantomsection
\label{sec:data_analysis}
\par
Analysis combined quantitative rates with physician-reviewed qualitative categories.

\subsubsection*{} % 2.8.1. Quantitative Analysis: Performance Metrics for Diagnostic Reliability
\textbf{2.8.1. Quantitative Analysis}

\textbf{Consistency Rate} measured the percentage of baseline scenarios for which a model produced clinically equivalent diagnoses across all consistency variants and repetitions:
\begin{equation}
    \text{Consistency Rate} = \left( \frac{\text{Number of Baseline Scenarios with Consistent Diagnoses}}{\text{Total Number of Baseline Scenarios}} \right) \times 100\%
\end{equation}

\textbf{Susceptibility Rate} measured the percentage of manipulated scenarios in which the model's diagnosis changed relative to the corresponding original prompt:
\begin{equation}
    \text{Susceptibility Rate} =
    \left( \frac{
    \begin{aligned}
        &\text{Number of Manipulated Scenarios} \\
        &\text{Resulting in a Changed Diagnosis}
    \end{aligned}
    }{\text{Total Number of Manipulated Scenarios}} \right) \times 100\%
\end{equation}

\textbf{Context Influence Rate} measured the percentage of contextually varied scenario sets in which the model changed its diagnosis across context-rich versions:
\begin{equation}
    \text{Context Influence Rate} =
    \left( \frac{
    \begin{aligned}
        &\text{Number of Contextually Varied Scenario Sets} \\
        &\text{Showing Diagnostic Changes}
    \end{aligned}
    }{
    \begin{aligned}
        &\text{Total Number of Contextually} \\
        &\text{Varied Scenario Sets}
    \end{aligned}
    } \right) \times 100\%
\end{equation}

A higher Context Influence Rate indicates greater diagnostic responsiveness but not necessarily better clinical reasoning; physician review was used to distinguish appropriate from inappropriate changes.

\subsubsection*{} % 2.8.2  Qualitative Analysis: Clinician-Informed Review of Diagnostic Changes
\textbf{2.8.2. Qualitative Analysis}

Qualitative categorisation followed the physician review protocol described in Section 2.7.7, with consensus categories integrated into the quantitative results.

\subsection*{} % 2.9. Ethical Considerations and Limitations
\textbf{2.9. Ethical Considerations and Limitations}
\phantomsection
\label{sec:ethical_considerations}

All cases were de novo synthetic scenarios, and no real Protected Health Information (PHI) was used.

Two limitations are noted. Demographic variation was part of the design, but the study was not powered for a bias or fairness audit across patient groups. Current LLMs also offer limited transparency into their reasoning, which constrains accountability in diagnostic settings.

\end{spacing}

\vspace{0.3cm}

%% file: results.tex
\textbf{III. Results} \vspace{-10pt} \\
\phantomsection
\label{sec:results}
\begin{spacing}{1.2}

\subsection*{} % Diagnostic Consistency: Perfect Clinical Equivalence Across Scenario Variations and Repeated Trials
\textbf{3.1. Diagnostic Consistency}\\
\vspace{-0.3cm}
\phantomsection
\label{sec:diagnostic_consistency}

Gemini and ChatGPT returned the same diagnosis for every baseline scenario, every clinically equivalent variant, and every repeated trial, achieving a \textbf{100.0\% baseline scenario consistency rate} across all 52 cases.

\begin{table}[h]
    \centering
    \begin{tabular}{lc}
        \toprule
        LLM & Baseline Scenario Consistency Rate (\%) \\
        \midrule
        Gemini & 100.0 \\
        ChatGPT & 100.0 \\
        \bottomrule
    \end{tabular}
    \caption{Baseline Scenario Consistency Rates of LLMs}
    \label{tab:consistency_rates}
\end{table}

\begin{center}
\begin{figure}[htbp]
    \centering
    \includegraphics[width=0.6\textwidth]{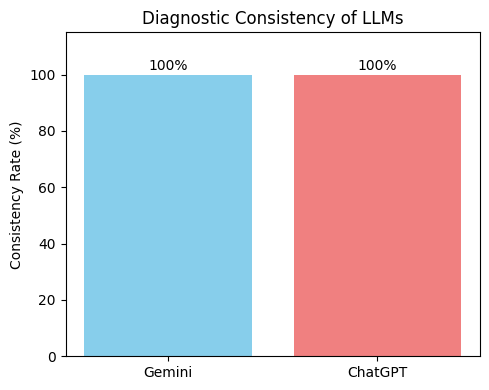}
    \caption{\fontsize{10pt}{20pt}\textit{Bar Chart Visualization of Baseline Scenario Consistency of LLMs}}
    \label{fig:consistency_bar_chart}
\end{figure}
\end{center}

Consistency under equivalent inputs is necessary but not sufficient: the same diagnosis every time does not mean the right diagnosis every time.

\subsection*{} % 2. Susceptibility to Manipulation: Differential Vulnerability to Diagnostic Shifts from Irrelevant Clinical Input
\textbf{3.2. Susceptibility to Manipulation}
\phantomsection
\label{sec:susceptibility_results}
\par
When irrelevant narrative details were added to prompts without changing the clinical evidence, Gemini shifted its diagnosis in 4 of 10 cases (\textbf{40.0\%}) and ChatGPT in 3 of 10 (\textbf{30.0\%}).

\begin{table}[h]
    \centering
    \begin{tabular}{lc}
        \toprule
        LLM & Cases with Changed Diagnosis (\%) \\
        \midrule
        Gemini & 40.0 \\
        ChatGPT & 30.0 \\
        \bottomrule
    \end{tabular}
    \caption{\fontsize{10pt}{20pt}\textit{Susceptibility Rates (Diagnosis Change) of LLMs Under Manipulation}}
    \label{tab:susceptibility_rates}
\end{table}

\begin{center}
\begin{figure}[h]
    \centering
    \includegraphics[width=0.7\textwidth]{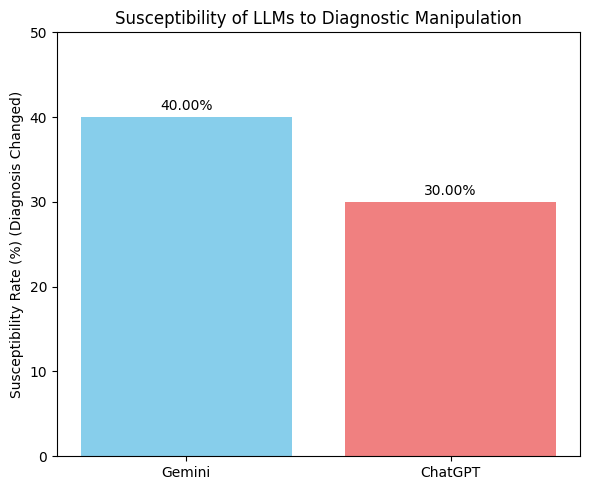}
    \caption{\fontsize{10pt}{20pt}\textit{Bar Chart Comparing Susceptibility Rates (Diagnosis Change) of LLMs}}
    \label{fig:susceptibility_bar_chart}
\end{figure}

\begin{figure}[h]
    \centering
    \includegraphics[width=0.7\textwidth]{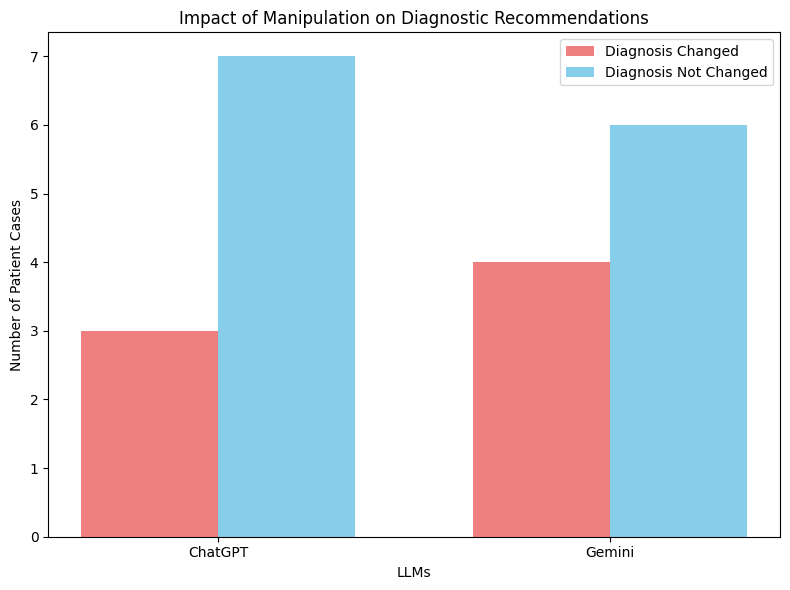}
    \caption{\fontsize{10pt}{20pt}\textit{Grouped Bar Chart of Diagnosis Change Status under Manipulation}}
    \label{fig:grouped_bar_chart}
\end{figure}
\end{center}

In each case, the clinical evidence was held constant; only the narrative framing changed. That either model shifted its diagnosis under these conditions is notable.

\subsection*{} % 3.3. Contextual Awareness
\textbf{3.3. Contextual Awareness}
\phantomsection
\label{sec:contextual_awareness_results}
\par
When clinically relevant context was added, ChatGPT changed its diagnosis in \textbf{77.8\%} of cases and Gemini in \textbf{55.6\%}. Whether those changes were clinically justified is addressed in the qualitative analysis below.

\begin{table}[h]
    \centering
    \begin{tabular}{lc}
        \toprule
        LLM & Context Influence Rate (\%) \\
        \midrule
        Gemini & 55.6 \\
        ChatGPT & 77.8 \\
        \bottomrule
    \end{tabular}
    \caption{\fontsize{10pt}{20pt}\textit{Context Influence Rates (Diagnosis Change due to Contextual Information) of LLMs}}
    \label{tab:context_influence_rates}
\end{table}

\begin{figure}[h]
    \centering
    \includegraphics[width=0.7\textwidth]{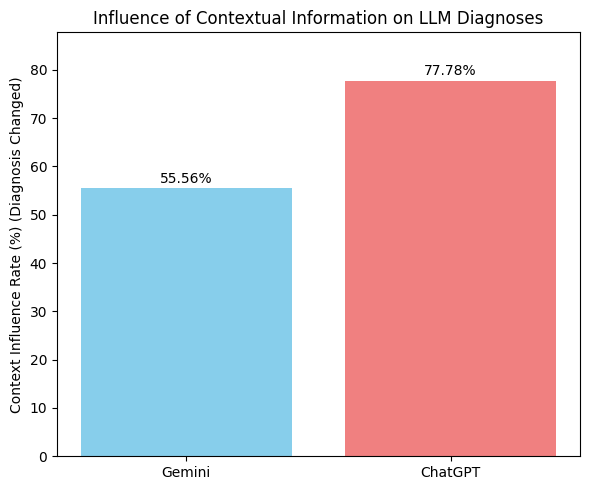}
    \caption{\fontsize{10pt}{20pt}\textit{Bar Chart Comparing Context Influence Rates (Diagnosis Change due to Context) of LLMs}}
    \label{fig:context_influence_bar_chart}
\end{figure}

\begin{figure}[h]
    \centering
    \includegraphics[width=0.7\textwidth]{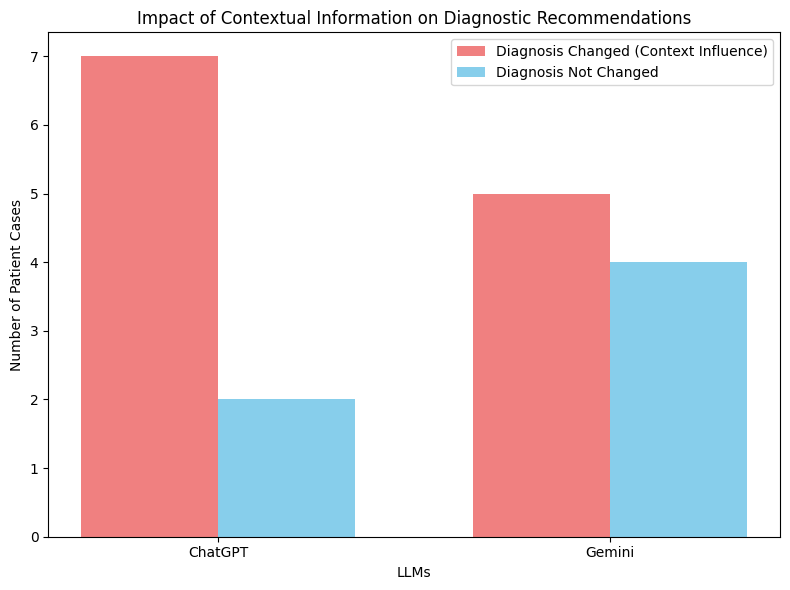}
    \caption{\fontsize{10pt}{20pt}\textit{Grouped Bar Chart of Diagnosis Change Status with Contextual Information}}
    \label{fig:grouped_context_bar_chart}
\end{figure}

\subsection*{} % 3.3.1. Contextual Awareness: Qualitative Analysis
\textbf{3.3.1. Contextual Awareness: Qualitative Analysis}

Two independent board-certified physicians reviewed context-driven diagnostic changes and categorised each as \textit{Appropriate}, \textit{Inappropriate}, or \textit{Ambiguous}. Inter-rater agreement was high (Cohen's $\kappa=0.85$), and final categories were resolved by consensus where needed.

\begin{table}[h]
    \centering
    \begin{tabular}{lcc}
        \toprule
        Category & Gemini (\%) & ChatGPT (\%) \\
        \midrule
        Appropriate Changes & 66.7 & 55.6 \\
        Inappropriate Changes & 22.2 & 33.3 \\
        Ambiguous Changes & 11.1 & 11.1 \\
        \bottomrule
    \end{tabular}
    \caption{\fontsize{10pt}{20pt}\textit{Consensus Categorization of Context-Driven Diagnostic Changes}}
    \label{tab:context_driven_changes}
\end{table}

\begin{table}[h]
    \centering
    \begin{tabular}{llccc}
        \toprule
        Model & Physician & Appropriate & Inappropriate & Ambiguous \\
        \midrule
        Gemini & X & 6 (66.67\%) & 2 (22.22\%) & 1 (11.11\%) \\
        Gemini & Y & 6 (66.67\%) & 2 (22.22\%) & 1 (11.11\%) \\
        ChatGPT & X & 5 (55.56\%) & 3 (33.33\%) & 1 (11.11\%) \\
        ChatGPT & Y & 5 (55.56\%) & 3 (33.33\%) & 1 (11.11\%) \\
        \bottomrule
    \end{tabular}
    \caption{\fontsize{10pt}{20pt}\textit{Physician-Specific Categorization (Counts/Percentages)}}
    \label{tab:physician_specific}
\end{table}

\begin{figure}[h]
    \centering
    \includegraphics[width=0.99\textwidth]{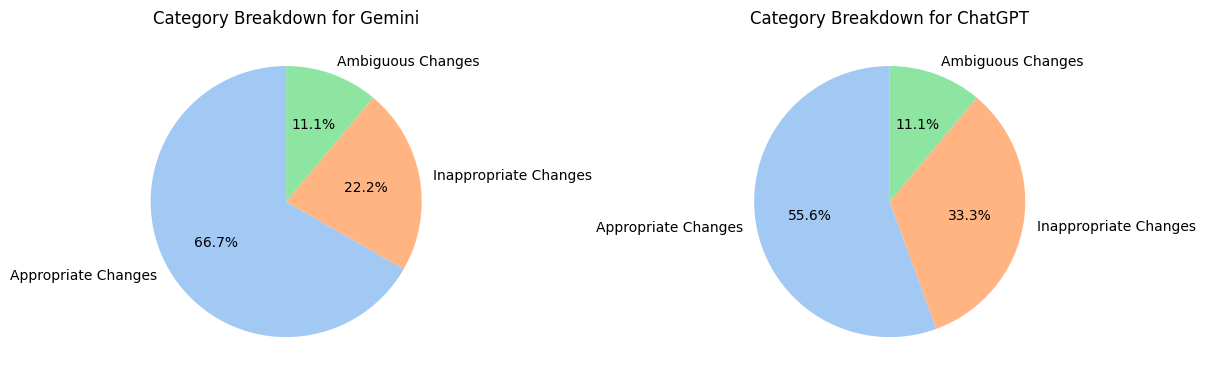}
    \caption{\fontsize{10pt}{20pt}\textit{Diagnosis Category Distribution per LLM - pie chart}}
    \label{fig:diagnosis_pie_chart}
\end{figure}

\subsection*{} % 3.3.2. Appropriate Context-Driven Changes
\textbf{3.3.2. Appropriate Context-Driven Changes}

Reviewers judged \textbf{66.7\%} of Gemini's context-driven shifts and \textbf{55.6\%} of ChatGPT's as clinically appropriate.

For instance, when prolonged fever and purulent nasal discharge were added to a viral upper-respiratory presentation, both models narrowed the diagnosis to \textit{\textbf{Bacterial Sinusitis}}.

\subsection*{} % 3.3.3. Inappropriate Context-Driven Changes
\textbf{3.3.3. Inappropriate Context-Driven Changes}

ChatGPT's context-driven shifts were judged inappropriate in \textbf{33.3\%} of reviewed cases, compared with \textbf{22.2\%} for Gemini.

\begin{figure}[h]
    \centering
    \includegraphics[width=0.7\textwidth]{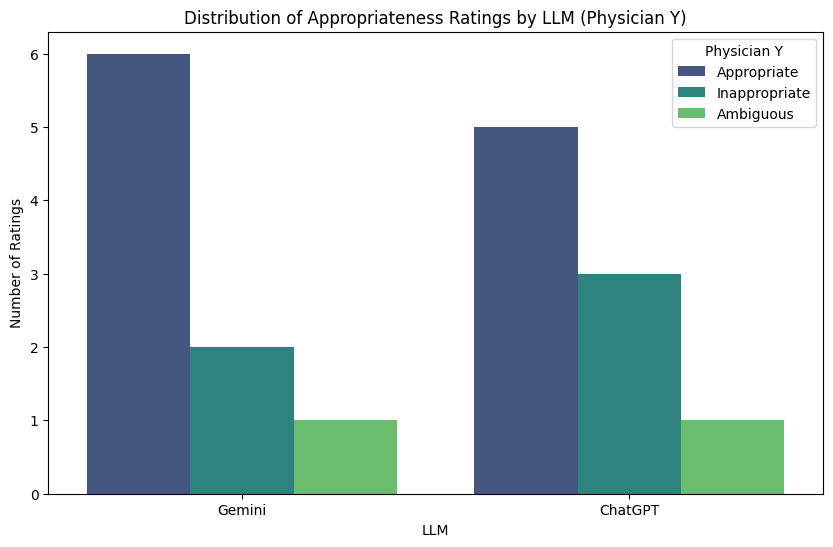}
    \caption{\fontsize{10pt}{20pt}\textit{Distribution of Appropriateness Ratings by LLM (Physician Y)}}
    \label{fig:appropriateness_physician_y}
\end{figure}

\begin{figure}[h]
    \centering
    \includegraphics[width=0.6\textwidth]{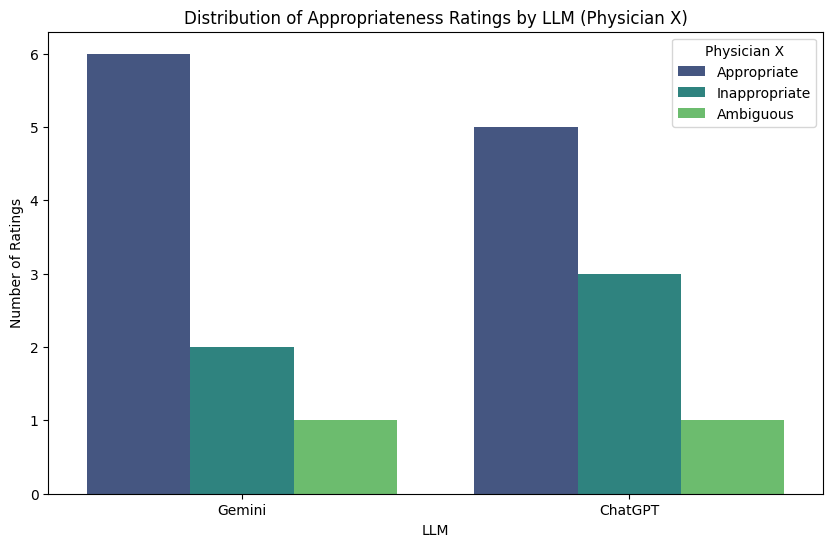}
    \caption{\fontsize{10pt}{20pt}\textit{Distribution of Appropriateness Ratings by LLM (Physician X)}}
    \label{fig:appropriateness_physician_x}
\end{figure}

One illustrative case involved Gemini changing an adult diagnosis from \textit{\textbf{Community-Acquired Pneumonia}} to \textit{\textbf{Bronchiolitis}} despite chest X-ray consolidation, which reviewers considered clinically inconsistent with the adult presentation and imaging findings.

\subsection*{} % 3.3.4. Ambiguous Context-Driven Changes
\textbf{3.3.4. Ambiguous Context-Driven Changes}

Reviewers classified \textbf{11.1\%} of context-driven shifts as ambiguous for both models.

\begin{figure}[h]
    \centering
    \includegraphics[width=0.66\textwidth]{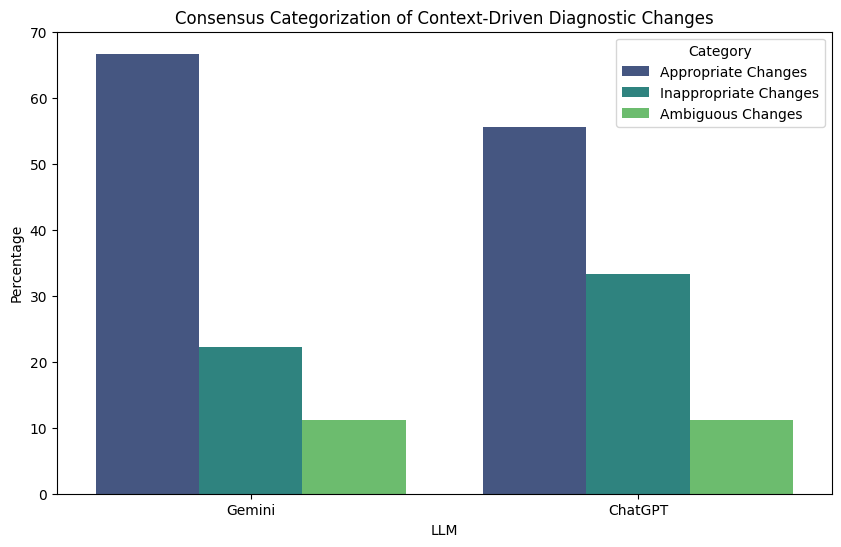}
    \caption{\fontsize{10pt}{20pt}\textit{Consistency Comparison: Physician Category Distribution by LLM}}
    \label{fig:consistency_physician_category}
\end{figure}

For example, Gemini's shift from \textit{\textbf{GERD}} to \textit{\textbf{Functional Dyspepsia}} was categorised as ambiguous because the scenario contained overlapping features and lacked additional clinical information, such as dietary history, response to acid suppression, or further diagnostic evaluation.

\begin{figure}[htbp]
    \centering
    \includegraphics[width=0.66 \textwidth]{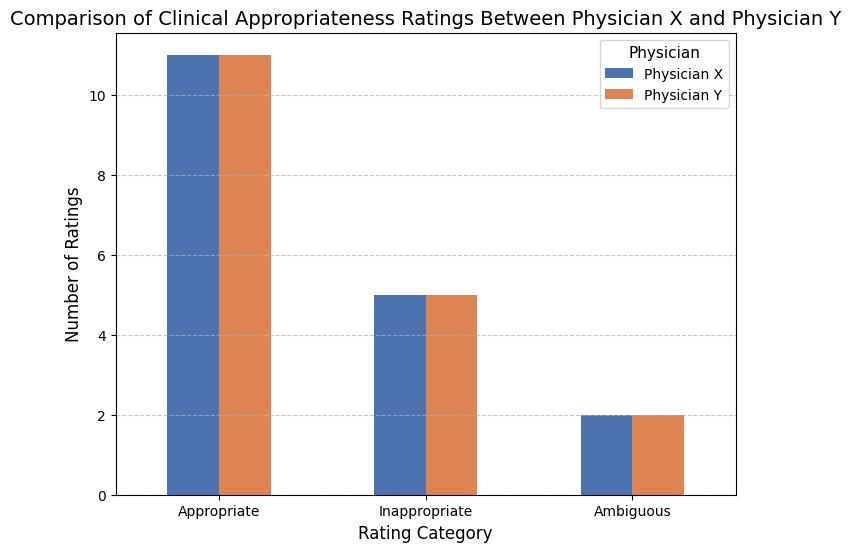}
    \caption{\fontsize{10pt}{20pt}\textit{Comparison of Physician X and Physician Y Ratings}}
    \label{fig:physician_ratings_comparison}
\end{figure}

\subsection*{} % 3.3.5. Differential Handling of Contextual Information
\textbf{3.3.5. Differential Handling of Contextual Information}
\par
\begin{itemize}
    \item \textbf{Better-handled contexts:} Both models incorporated biomarker trends and imaging findings appropriately. Rising creatinine raised concern for acute kidney injury; CT evidence of pulmonary embolism shifted diagnoses toward PE.
    \item \textbf{Less reliably handled contexts:} Social-history details and nonspecific vital-sign changes were over-weighted. Introducing homelessness triggered shifts toward substance-related psychosis; isolated tachycardia was treated as evidence for sepsis without supporting findings.
\end{itemize}

\subsection*{} % 3.3.6. Balancing Diagnostic Responsiveness and Clinical Soundness
\vspace{0.3cm}
\textbf{3.3.6. Balancing Diagnostic Responsiveness and Clinical Soundness}
\par
Responding to context more often did not mean responding better. ChatGPT changed diagnoses more frequently, but a larger share of its changes were inappropriate; Gemini changed less often, but with a higher rate of clinically justified shifts.

\begin{figure}[h]
    \centering
    \includegraphics[width=0.7\textwidth]{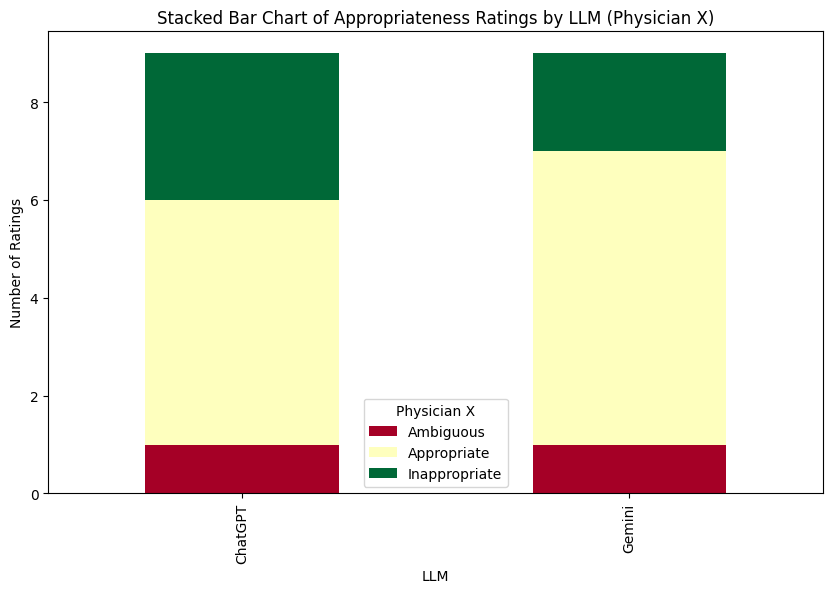}
    \caption{\fontsize{10pt}{20pt}\textit{Stacked Bar Chart of Appropriateness Ratings by LLM (Physician X)}}
    \label{fig:stacked_appropriateness_physician_x}
\end{figure}

\begin{figure}[h]
    \centering
    \includegraphics[width=0.7\textwidth]{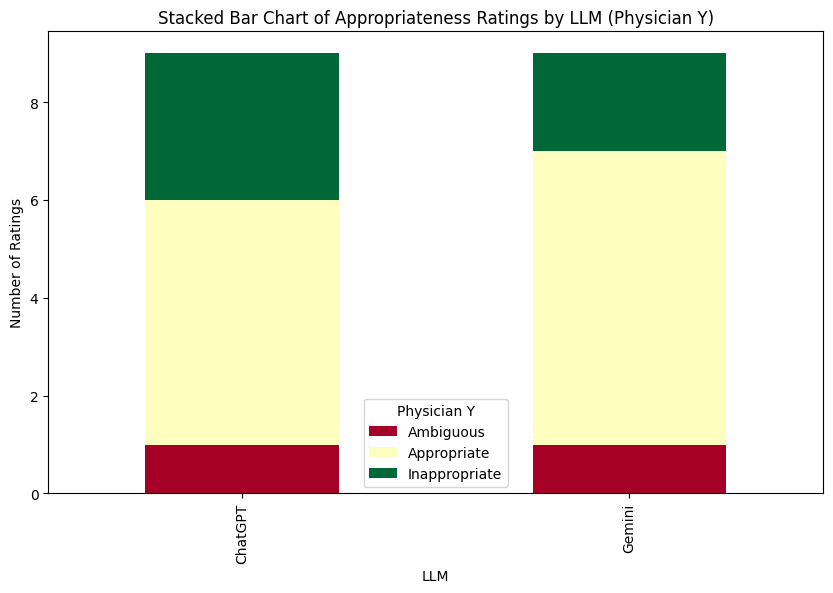}
    \caption{\fontsize{10pt}{20pt}\textit{Stacked Bar Chart of Appropriateness Ratings by LLM (Physician Y)}}
    \label{fig:stacked_appropriateness_physician_y}
\end{figure}

\end{spacing}

%% file: discussion.tex
\textbf{IV. Discussion} \vspace{-10pt} \\
\phantomsection
\label{sec:discussion}

\begin{spacing}{1.2}
Gemini and ChatGPT both showed stable diagnoses under controlled, clinically equivalent inputs. That stability did not extend to prompts with irrelevant content or to cases where clinical context shifted. The three experiments revealed that reproducible output and sound clinical reasoning are not the same thing.

\subsubsection*{} % 4.1 Diagnostic Consistency:
\textbf{4.1. Diagnostic Consistency}
\phantomsection
\label{sec:discussion_consistency}
\par
Both models achieved \textbf{100\% diagnostic consistency} across the 52 baseline scenarios and their clinically equivalent variants. In a diagnostic support role, this reproducibility matters: the same case presented twice should not yield different answers~\cite{Shortliffe2018}.

But consistency is not accuracy~\cite{Ghassemi2021, Marcus2020}. If a model overweights a misleading phrase, misses absent information, or carries bias from training data, high consistency locks that error in across every query~\cite{Amodei2016}. The 100\% consistency rate tells us these models are stable under equivalent inputs. It tells us nothing about whether their diagnoses are correct.

\subsubsection*{} % 4.2. Susceptibility to Manipulation: Differential Vulnerability and Clinically Significant Diagnostic Shifts
\textbf{4.2. Susceptibility to Manipulation}
\phantomsection
\label{sec:discussion_susceptibility}
\par
Irrelevant but plausible prompt additions changed Gemini's diagnosis in \textbf{40.0\%} of manipulated scenarios and ChatGPT's in \textbf{30.0\%}. The core diagnostic evidence was unchanged in every case, so these shifts reflect sensitivity to content that had no bearing on the correct diagnosis~\cite{Finlayson2019, Goodfellow2015, Szegedy2014}.

Patient narratives routinely contain irrelevant details, subjective interpretations, and incidental history~\cite{Char2018, Mittelstadt2016}. A model that shifts its diagnosis in response to such content is unreliable for any case where the prompt is not pre-cleaned. The shifts we observed were not trivial: they included movement from a high-acuity cardiac diagnosis to a less urgent one and from one cardiopulmonary condition to another with different management paths.

These models process whatever they receive. They do not question whether a detail is relevant, whether evidence conflicts, or whether critical data is missing~\cite{Marcus2020}. Clinicians are also susceptible to framing effects, but clinical training builds in checkpoints: reassessment, examination, targeted testing~\cite{Kawamoto2005, Shortliffe2018}. LLMs have no equivalent checkpoint unless one is imposed externally.

\begin{longtable}{p{3cm} p{4cm} p{4cm} p{4cm}}
    \caption{Comparison of LLM and Human Clinician Diagnostic Factors} \label{tab:diagnostic_factors} \\
    \toprule
    \textbf{Factor} & \textbf{LLM Weaknesses} & \textbf{Human Clinician Strengths} & \textbf{Human Clinician Weaknesses} \\
    \midrule
    \endfirsthead

    \caption[]{Comparison of LLM and Human Clinician Diagnostic Factors (Continued)} \\
    \toprule
    \textbf{Factor} & \textbf{LLM Weaknesses} & \textbf{Human Clinician Strengths} & \textbf{Human Clinician Weaknesses} \\
    \midrule
    \endhead

    \endfoot

    \bottomrule
    \endlastfoot

    \textbf{Anchoring Bias} & Susceptible to leading or anchoring prompt details. & Uses clinical protocols and reassessment to reduce anchoring. & Can still anchor on early or salient information. \\
    \addlinespace
    \textbf{Contextual Gaps} & Struggles with missing, conflicting, or irrelevant data. & Actively seeks missing history, examination findings, and tests. & May make assumptions when information is incomplete. \\
    \addlinespace
    \textbf{Irrelevant Information} & May overweight irrelevant prompt content. & Filters information using clinical relevance and judgment. & Can be distracted by salient but irrelevant findings. \\
    \addlinespace
    \textbf{Diagnostic Certainty} & May provide a final diagnosis without sufficient evidence. & Can express uncertainty and defer diagnosis pending further data. & May show overconfidence or premature closure. \\
    \addlinespace
    \textbf{Input Validity} & Does not reliably question input validity. & Can question inconsistencies and seek corroboration. & May be influenced by source credibility or framing. \\
\end{longtable}

In clinical use, LLM diagnostic output should be checked against objective findings and not treated as final when the input has not been vetted for irrelevant or conflicting content.

\subsubsection*{} % 4.3. Contextual Awareness: Differential Responsiveness, Clinical Appropriateness, and Data Handling Limitations
\textbf{4.3. Contextual Awareness}
\phantomsection
\label{sec:discussion_contextual}
\par
Changing a diagnosis in response to new context is not the same as reasoning well about that context. ChatGPT changed its diagnosis more often than Gemini when context was added (\textbf{77.8\%} vs. \textbf{55.6\%}), but physician review rated a smaller share of ChatGPT's changes as appropriate (\textbf{55.6\%} vs. \textbf{66.7\%} for Gemini) and a larger share as inappropriate (\textbf{33.3\%} vs. \textbf{22.2\%}).

\begin{longtable}{p{3.5cm} c c p{4.5cm}}
    \caption{Quantitative Context Influence vs. Qualitative Appropriateness} \label{tab:context_influence} \\
    \toprule
    \addlinespace
    \textbf{Feature} & \makecell{\textbf{Gemini} \\ \textbf{(55.6\% CIR)}} & \makecell{\textbf{ChatGPT} \\ \textbf{(77.8\% CIR)}} & \textbf{Interpretation} \\ [0.9em]
    \midrule
    \endfirsthead

    \caption[]{Quantitative Context Influence vs. Qualitative Appropriateness (Continued)} \\
    \toprule
    \addlinespace
    \textbf{Feature} & \makecell{\textbf{Gemini} \\ \textbf{(55.6\% CIR)}} & \makecell{\textbf{ChatGPT} \\ \textbf{(77.8\% CIR)}} & \textbf{Interpretation} \\ [0.9em]
    \midrule
    \endhead

    \endfoot

    \bottomrule
    \endlastfoot

    \addlinespace
    \parbox[t]{3.5cm}{\textbf{Context\\Influence Rate}} & 55.6\% & 77.8\% & ChatGPT changed diagnoses more often after context was added. \\ [0.5cm]
    \addlinespace
    \parbox[t]{3.5cm}{\textbf{Appropriate\\Changes}} & 66.7\% & 55.6\% & Gemini had a higher proportion of clinically justified context-driven changes. \\ [1.2cm]
    \addlinespace
    \parbox[t]{3.5cm}{\textbf{Inappropriate\\Changes}} & 22.2\% & 33.3\% & ChatGPT had a higher proportion of clinically unjustified context-driven shifts. \\ [1.2cm]
\end{longtable}

When context provided strong evidence (ECG changes, gastrointestinal bleeding), both models refined their diagnoses appropriately~\cite{Amann2020, Arrieta2020}. When the evidence was weaker (remote history, vague symptoms, incomplete data), both also made unjustified shifts~\cite{Ghassemi2021, Holzinger2017}.

\begin{longtable}{p{1.8cm}p{1.6cm}p{2cm}p{3.6cm}p{2.7cm}p{3cm}}
    \caption{Examples of Clinically Appropriate Context-Driven Diagnostic Refinements} \label{tab:appropriate_refinements} \\
    \toprule
    \addlinespace
    \textbf{LLM} & \textbf{Case ID} & \textbf{Original Diagnosis} & \textbf{Contextual Information Added} & \textbf{Context-Driven Diagnosis Refinement} & \textbf{Clinical Benefit} \\
    \midrule
    \endfirsthead

    \caption[]{Examples of Clinically Appropriate Context-Driven Diagnostic Refinements (Continued)} \\
    \toprule
    \addlinespace
    \textbf{LLM} & \textbf{Case ID} & \textbf{Original Diagnosis} & \textbf{Contextual Information Added} & \textbf{Context-Driven Diagnosis Refinement} & \textbf{Clinical Benefit} \\
    \midrule
    \endhead

    \endfoot

    \bottomrule
    \endlastfoot

    \addlinespace
    \textbf{Gemini} & \textbf{001\_3\_2} & Angina Pectoris & ECG findings suggestive of instability & Unstable Angina & More appropriate risk stratification. \\
    \addlinespace
    \textbf{ChatGPT} & \textbf{024\_1} & Peptic Ulcer Disease & Gastrointestinal bleeding context & Gastric Ulcer with Gastrointestinal Bleeding & Improved diagnostic specificity. \\
    \addlinespace
    \textbf{Gemini / ChatGPT} & \textbf{Multiple} & Broad initial diagnosis & Lifestyle or demographic details & More specific subtype or qualifier & More individualized diagnostic framing. \\
    \addlinespace
\end{longtable}

\begin{longtable}{p{1.8cm}p{1.7cm}>{\raggedright\arraybackslash}p{2.0cm}>{\raggedright\arraybackslash}p{2.7cm}>{\raggedright\arraybackslash}p{2.5cm}>{\raggedright\arraybackslash}p{3.8cm}}
\caption{Examples of Clinically Inappropriate Shifts and Diagnostic Overconfidence} \label{tab:inappropriate_shifts} \\
\toprule
\textbf{LLM} & \textbf{Case ID} & \textbf{Original Diagnosis} & \textbf{Contextual Information Added} & \textbf{Context-Driven Diagnosis Shift} & \textbf{Clinical Concern} \\
\midrule
\endfirsthead
\toprule
\textbf{LLM} & \textbf{Case ID} & \textbf{Original Diagnosis} & \textbf{Contextual Information Added} & \textbf{Context-Driven Diagnosis Shift} & \textbf{Clinical Concern} \\
\midrule
\endhead
\midrule
\endfoot
\bottomrule
\endlastfoot
\small
\textbf{Gemini} & \textbf{001\_3\_3} & Atypical\newline MI\newline (NSTEMI) & Recent history of ``heavy lifting'' & Musculo-\newline skeletal\newline Back Pain & Shifted away from a cardiac diagnosis despite cardiac markers. \\
\addlinespace
\textbf{ChatGPT} & \textbf{024\_4} & Acute\newline Bronchitis & Childhood asthma history & Asthma\newline Exacerbation & Overweighted remote history and underweighted the acute infectious presentation. \\
\addlinespace
\textbf{Gemini} & \textbf{024\_3} & Peptic\newline Ulcer\newline Disease & Vague symptoms, no confirmatory tests & Possible\newline Malignancy\newline (Gastric\newline Cancer) & Suggested a serious diagnosis without sufficient confirmatory evidence. \\
\addlinespace
\textbf{Both} & \textbf{Multiple} & Diagnosis\newline requiring\newline test\newline results & Scenarios with incomplete data & Diagnosis\newline without\newline requesting\newline tests & Produced final diagnoses despite missing information. \\
\end{longtable}

The core issue is how evidence is prioritised. Clinicians rank objective findings, red flags, temporal patterns, and disease prevalence before revising a working diagnosis~\cite{Shortliffe2018, Osheroff2012}. Neither model showed this kind of hierarchical reasoning~\cite{Obermeyer2019}.

\begin{longtable}{p{3.5cm}p{6cm}p{6cm}}
    \caption{Comparison of LLMs and Human Clinicians in Contextual Reasoning}
    \label{tab:contextual_reasoning} \\
    \toprule
    \textbf{Aspect} & \textbf{LLMs} & \textbf{Human Clinicians} \\
    \midrule
    \endfirsthead

    \caption[]{Comparison of LLMs and Human Clinicians in Contextual Reasoning (Continued)} \\
    \toprule
    \textbf{Aspect} & \textbf{LLMs} & \textbf{Human Clinicians} \\
    \midrule
    \endhead

    \bottomrule
    \endlastfoot

    \textbf{Use of Context} & Can change diagnoses when new information is added, but may overweight weak cues. & Integrates context with clinical probability, examination, tests, and patient history. \\
    \addlinespace
    \textbf{Handling Missing Data} & May provide a final diagnosis despite missing essential information. & Can defer diagnosis and request additional history, examination, or tests. \\
    \addlinespace
    \textbf{Noise Filtering} & May treat irrelevant details as diagnostically meaningful. & Filters irrelevant details through clinical judgment. \\
    \addlinespace
    \textbf{Diagnostic Restraint} & May sound definitive even when evidence is incomplete. & Can express uncertainty and maintain differential diagnoses. \\
    \addlinespace
    \textbf{Clinical Accountability} & Does not bear responsibility for patient outcomes. & Responsible for clinical decisions, communication, and follow-up. \\
\end{longtable}

\subsection*{}
\textbf{4.4. Synthesis}
\phantomsection
\label{sec:integrated_discussion}
\par
The three experiments expose a pattern we call the \textbf{fragility triad}. First, \textbf{deterministic consistency}: both models reproduce the same output under equivalent inputs, but that same determinism can lock in systematic errors. Second, \textbf{input blindness}: neither model distinguishes diagnostically relevant content from irrelevant additions~\cite{Finlayson2019, Goodfellow2015}. Third, \textbf{contextual reasoning gaps}: added context can improve a diagnosis, but it can also trigger unjustified shifts when the cue is weak or the data incomplete~\cite{Ghassemi2021, Obermeyer2019}.

A diagnostic tool must do more than respond to information. It must decide which information matters, how much weight it deserves, and when the available evidence is not enough to support a conclusion~\cite{Floridi2018, Floridi2019, Amodei2016, Mittelstadt2016}.

\subsubsection*{} % Implications for Clinical Use
\textbf{4.5. Implications for Clinical Use}
\phantomsection
\label{sec:implications_clinical}
\par
These results support using LLMs as assistive tools, not autonomous diagnosticians~\cite{Topol2019, Beam2018, Rajpurkar2022}. Their consistency is most useful in structured, low-risk tasks (terminology standardization, case summarisation, preliminary differentials) where a clinician reviews every output~\cite{Shortliffe2018, Osheroff2012}.

The risks grow in complex or high-acuity presentations. Irrelevant narrative details can shift a diagnosis. Weak context can push it away from the best-supported interpretation. And a model that never asks for missing tests or expresses uncertainty can encourage premature closure if its output goes unchecked.

Responsible clinical integration requires several safeguards~\cite{FDA2021, Benjamens2020, WorldHealthOrganization2021, Beauchamp2019, GDPR2018}:
\begin{itemize}
    \item \textbf{Clinician oversight:} LLM outputs must be reviewed by a clinician before use in patient care.
    \item \textbf{Input validation:} systems should flag irrelevant, conflicting, or incomplete information before generating diagnostic recommendations.
    \item \textbf{Uncertainty handling:} models should state when evidence is insufficient and what additional data are needed.
    \item \textbf{Output verification:} diagnostic suggestions should be checked against objective findings, guideline-based criteria, and clinical plausibility.
    \item \textbf{Defined use cases:} deployment should be limited to settings where risks, oversight responsibilities, and escalation pathways are explicit.
\end{itemize}

\subsubsection*{} % Limitations and Future Work
\textbf{4.6. Limitations and Future Work}
\phantomsection
\label{sec:limitations_future}
\par
This study has several limitations~\cite{NationalAcademyofMedicine2022, AINowInstitute2023}. The 52 scenarios were synthetic; they allow experimental control but do not capture the ambiguity, multimorbidity, or workflow constraints of real clinical encounters. Manipulation testing covered ten scenarios with specific categories of irrelevant text; broader adversarial patterns remain untested~\cite{Finlayson2019}.

Contextual-awareness testing drew on nine scenarios from two baseline cases; enough for focused analysis, not enough to generalise across clinical domains. The evaluation covered only Gemini 2.0 Flash and ChatGPT-4o as accessed in February 2025; commercial models change frequently, and results may not apply to later versions or domain-specific medical LLMs. Finally, this study tested reliability, not diagnostic accuracy or patient outcomes.

Future work should expand the case set, include real clinical data where ethics allow, test more models, and track whether these failures persist across model updates~\cite{Bommasani2021}. Comparing LLM output against expert consensus and patient outcomes would clarify clinical relevance. On the engineering side, adversarial training~\cite{Goodfellow2015}, structured input formats, uncertainty calibration, and interfaces that require confirmation of missing data before producing a diagnosis~\cite{Bucinca2021} are all worth testing.

Gemini and ChatGPT are reproducible and sometimes refine their diagnoses well when given relevant context~\cite{Char2018, Floridi2018, WorldHealthOrganization2021}. They are also influenced by content that should not matter and produce unjustified diagnostic shifts when context is weak. Unless these failure modes are resolved, any clinical use of these models demands human oversight at every step.

\end{spacing}

%% file: conclusion.tex
\subsubsection*{} % VII. Conclusion
\textbf{V. Conclusion} \vspace{-10pt} \\
\phantomsection
\label{sec:conclusion1}
\begin{spacing}{1.2}

Both models were reproducible under controlled conditions, but reproducibility alone did not establish clinical reliability~\cite{Amodei2016, Marcus2020}.

Both achieved \textbf{100\% consistency} across repeated and varied presentations. Yet irrelevant prompt additions shifted diagnoses in \textbf{40.0\%} of Gemini cases and \textbf{30.0\%} of ChatGPT cases; the models were stable when nothing changed but vulnerable when noise was introduced.

Responsiveness to context did not track with clinical soundness. ChatGPT changed diagnoses more often (\textbf{77.8\%} vs. \textbf{55.6\%}), but more of its changes were inappropriate (\textbf{33.3\%} vs. \textbf{22.2\%}); Gemini changed less often but with a higher appropriate-change rate (\textbf{66.7\%} vs. \textbf{55.6\%}).

These models can reproduce results and, when given strong context, refine diagnoses appropriately. But they also shifted diagnoses in response to irrelevant input, made unjustified changes on weak evidence, and never flagged missing data. They are not autonomous diagnostic tools.

Their value lies in structured, supervised roles: case summarisation, preliminary differentials, terminology standardisation, where a clinician reviews every output~\cite{Topol2019, Shortliffe2018}. Clinical deployment requires input validation, uncertainty signalling, and human review at every decision point. Future work should expand the case set, test additional models, and evaluate whether adversarial training or structured input formats reduce the failure modes observed here~\cite{Bommasani2021, NationalAcademyofMedicine2022}.

\end{spacing}

%% file: appendix.tex
\subsection*{}
\textbf{\normalfont\large\bfseries Appendix A: Data Availability} \vspace{-10pt} \\
\phantomsection
\label{sec:app1}

\begin{spacing}{1.2}

The clinical scenarios, LLM interactions, and aggregated results are publicly available in a GitHub repository:

\url{https://github.com/neryva-lab/The-Reliability-of-LLMs-for-Medical-Diagnosis-Data.git}
\vspace{10pt}

The repository contains:

\begin{itemize}[leftmargin=*]
    \item \textbf{Clinical scenario datasets (JSON):} Baseline and modified clinical scenarios used for the evaluations.
    \item \textbf{LLM responses (JSON):} Raw responses from Gemini and ChatGPT.
    \item \textbf{Aggregated results (JSON and CSV):} Summary files for the consistency, manipulation, and contextual-awareness analyses.
    \item \textbf{Prompt templates (text):} Standardized prompt templates used for LLM interactions.
    \item \textbf{README file:} Repository contents, data formats, and usage notes.
\end{itemize}

\subsection*{}
\textbf{\normalfont\large\bfseries Appendix B: Example Baseline Scenario} \\
\phantomsection
\label{sec:app2}
\vspace{-0.3cm}
\par
The following example illustrates the structure of a baseline clinical scenario before experimental modifications.

\subsubsection*{}
\textbf{Example Baseline Scenario: Case 001 - Acute Coronary Syndrome}
\subsubsection*{}
\textbf{\textit{Patient Demographics and History:}}

\begin{itemize}[leftmargin=*]
    \item \textbf{Patient ID:} 001
    \item \textbf{Age:} 65 years
    \item \textbf{Gender:} Male
    \item \textbf{Medical History:}
    \begin{itemize}[leftmargin=*,noitemsep,topsep=0pt]
        \item[$\circ$] Hypertension
        \item[$\circ$] Type 2 Diabetes
        \item[$\circ$] Previous Myocardial Infarction (MI)
    \end{itemize}
    \item \textbf{Current Medications:}
    \begin{itemize}[leftmargin=*,noitemsep,topsep=0pt]
        \item[$\circ$] Aspirin
        \item[$\circ$] Lisinopril
        \item[$\circ$] Metformin
    \end{itemize}
\end{itemize}

\subsubsection*{}
\textbf{\textit{Presenting Complaint and Symptoms:}}

\begin{itemize}[leftmargin=*]
    \item \textbf{Presenting Complaint:} Chest pain
    \item \textbf{Symptoms:}
    \begin{itemize}[leftmargin=*,noitemsep,topsep=0pt]
        \item[$\circ$] \textbf{Chest Pain:}
        \begin{itemize}[leftmargin=*,noitemsep,topsep=0pt]
            \item Severity: Moderate
            \item Duration: 30 minutes
            \item Location: Retrosternal
            \item Character: Crushing
            \item Associated Symptoms: Radiating to left arm, diaphoresis
        \end{itemize}
        \item[$\circ$] \textbf{Shortness of Breath:}
        \begin{itemize}[leftmargin=*,noitemsep,topsep=0pt]
            \item Severity: Mild
            \item Duration: Intermittent
            \item Exacerbating Factors: Exertion
            \item Relieving Factors: Rest
        \end{itemize}
        \item[$\circ$] \textbf{Nausea:}
        \begin{itemize}[leftmargin=*,noitemsep,topsep=0pt]
            \item Severity: Mild
            \item Duration: Intermittent
        \end{itemize}
        \item[$\circ$] \textbf{Diaphoresis:}
        \begin{itemize}[leftmargin=*,noitemsep,topsep=0pt]
            \item Severity: Mild
            \item Duration: Intermittent
            \item Type: Cold sweat
        \end{itemize}
    \end{itemize}
\end{itemize}

\subsubsection*{}
\textbf{\textit{Vital Signs:}}

\begin{itemize}[leftmargin=*]
    \item \textbf{Heart Rate:} 100 bpm
    \item \textbf{Blood Pressure:} 150/90 mmHg
    \item \textbf{Temperature:} 98.6 \textdegree F (37 \textdegree C)
    \item \textbf{Respiratory Rate:} 20 breaths/min
\end{itemize}

\subsubsection*{}
\textbf{\textit{Physical Exam Findings:}}

\begin{itemize}[leftmargin=*]
    \item Patient appears anxious.
    \item No murmurs, rubs, or gallops auscultated.
    \item Lungs clear to auscultation.
\end{itemize}

\subsubsection*{}
\textbf{\textit{Test Results:}}

\begin{itemize}[leftmargin=*]
    \item \textbf{ECG:} ST elevation in leads II, III, aVF
    \item \textbf{Troponin:} 0.8 ng/mL (Elevated)
    \item \textbf{CBC:} White blood cell (WBC) count elevated
\end{itemize}

\subsubsection*{}
\textbf{\textit{Diagnosis and Management (for reference in scenario design, not provided to LLMs):}}

\begin{itemize}[leftmargin=*]
    \item \textbf{Differential Diagnosis:}
    \begin{itemize}[leftmargin=*,noitemsep,topsep=0pt]
        \item[$\circ$] Myocardial Infarction
        \item[$\circ$] Angina Pectoris
        \item[$\circ$] Pericarditis
        \item[$\circ$] Pulmonary Embolism
    \end{itemize}
    \item \textbf{Final Diagnosis:} Myocardial Infarction (STEMI)
    \item \textbf{Treatment:} Aspirin, Clopidogrel, Thrombolytic therapy
    \item \textbf{Notes:} Patient responded well to thrombolytic therapy. Transferred to CCU for further monitoring.
\end{itemize}

\section*{}
\textbf{\normalfont\large\bfseries Appendix C: Prompt Sample} \vspace{-10pt} \\
\phantomsection
\label{sec:app3}

This appendix shows the prompt structure used to elicit diagnostic outputs from both models.

\subsection*{}

Example prompt:

\begin{jsonbox}
You are a medical doctor. Diagnose the following patient case.\\
\end{jsonbox}

\vspace{-1.2cm}

\begin{jsonbox}
[\\
  {\\
    \jsonkey{patient\_id}: \jsonstring{001\_1},\\
    \jsonkey{age}: \jsonnumber{65},\\
    \jsonkey{gender}: \jsonstring{Male},\\
    \jsonkey{country}: \jsonstring{USA},\\
    \jsonkey{race}: \jsonstring{White},\\
    \jsonkey{ethnicity}: \jsonstring{Not Hispanic or Latino},\\
    \jsonkey{medical\_history}: [\jsonstring{Hypertension}, \jsonstring{Type 2 Diabetes}, \jsonstring{History of MI}],\\
    \jsonkey{current\_medications}: [\jsonstring{Aspirin}, \jsonstring{Lisinopril}, \jsonstring{Metformin}],\\
    \jsonkey{presenting\_complaint}: \jsonstring{Chest discomfort},\\
    \jsonkey{symptoms}: [\\
      {\\
        \jsonkey{name}: \jsonstring{Chest pain},\\
        \jsonkey{severity}: \jsonstring{moderate},\\
        \jsonkey{duration}: \jsonstring{30 minutes},\\
        \jsonkey{location}: \jsonstring{retrosternal},\\
        \jsonkey{character}: \jsonstring{crushing},\\
        \jsonkey{associated\_symptoms}: [\jsonstring{radiating to left arm}, \jsonstring{diaphoresis}]\\
      },\\
      {\\
        \jsonkey{name}: \jsonstring{Shortness of breath},\\
        \jsonkey{severity}: \jsonstring{mild},\\
        \jsonkey{duration}: \jsonstring{intermittent},\\
        \jsonkey{exacerbating\_factors}: \jsonstring{exertion},\\
        \jsonkey{relieving\_factors}: \jsonstring{rest}\\
      }\\
    ],\\
    \jsonkey{vital\_signs}: {\\
      \jsonkey{heart\_rate}: \jsonnumber{100},\\
      \jsonkey{blood\_pressure}: \jsonstring{150/90},\\
      \jsonkey{temperature}: \jsonnumber{98.6},\\
      \jsonkey{respiratory\_rate}: \jsonnumber{20}\\
    },\\
    \jsonkey{physical\_exam}: \jsonstring{Patient appears anxious. Cardiovascular exam unremarkable. Lungs clear.},\\
    \jsonkey{test\_results}: {\\
      \jsonkey{ecg}: \jsonstring{ST elevation in leads II, III, aVF},\\
      \jsonkey{troponin}: \jsonnumber{0.8},\\
      \jsonkey{cbc}: \jsonstring{WBC count elevated}\\
    },\\
    \jsonkey{differential\_diagnosis}: [],\\
    \jsonkey{final\_diagnosis}: \jsonstring{},\\
    \jsonkey{treatment}: \jsonstring{},\\
    \jsonkey{notes}: \jsonstring{}\\
  }\\
]\\
\end{jsonbox}

\vspace{-1.2cm}

\begin{jsonbox}
Provide patient\_id and corresponding diagnosis for that patient.\\
Provide ONLY your final diagnosis.\\
Do not provide explanations or differential diagnoses.\\
\end{jsonbox}

The prompt contains four elements:

\begin{itemize}
    \item \textbf{Role instruction:} \textit{You are a medical doctor.}
    \item \textbf{Task definition:} \textit{Diagnose the following patient case.}
    \item \textbf{JSON input:} Patient details are provided in a structured format, including demographics, history, symptoms, vital signs, and test results.
    \item \textbf{Output constraints:}
    \begin{itemize}
        \item[$\circ$] Provide \texttt{patient\_id} and the final diagnosis.
        \item[$\circ$] Provide \textbf{ONLY} the final diagnosis, without explanations or differential diagnoses.
    \end{itemize}
\end{itemize}

\end{spacing}